\documentclass{article}

\usepackage[final,nonatbib]{neurips_2024}

\usepackage[utf8]{inputenc} % allow utf-8 input
\usepackage[T1]{fontenc}    % use 8-bit T1 fonts
\usepackage{hyperref}       % hyperlinks
\usepackage{url}            % simple URL typesetting
\usepackage{booktabs}       % professional-quality tables
\usepackage{amsfonts}       % blackboard math symbols
\usepackage{nicefrac}       % compact symbols for 1/2, etc.
\usepackage{microtype}      % microtypography
\usepackage{subcaption}
\usepackage{graphicx}
\usepackage{color, colortbl}
 \usepackage{amsmath}
 \usepackage{amsfonts}
 \usepackage{multirow}
 \usepackage{wrapfig}

\usepackage{color, colortbl}
\definecolor{Gray}{gray}{0.9}
\definecolor{TextGreen}{rgb}{0.0,0.69,0.31}
\definecolor{babyblue}{rgb}{0.54, 0.81, 0.94}
\definecolor{firebrick}{rgb}{0.7, 0.13, 0.13}
\definecolor{flame}{rgb}{0.89, 0.35, 0.13}

\usepackage{algorithm}
\usepackage{algorithmic}

\usepackage[dvipsnames]{xcolor}

\definecolor{Gray}{gray}{0.9}

\DeclareMathOperator*{\argmax}{arg\,max}

\title{
Open-Vocabulary Object Detection via Language Hierarchy
}

\author{
  Jiaxing Huang, Jingyi Zhang, Kai Jiang, Shijian Lu\thanks{Corresponding author}\\
  College of Computing and Data Science \\
  Nanyang Technological University, Singapore\\
}

\begin{document}

\maketitle

\begin{abstract}
Recent studies on generalizable object detection have attracted increasing attention with additional weak supervision from large-scale datasets with image-level labels.
However, weakly-supervised detection learning often suffers from image-to-box label mismatch, i.e., image-level
labels do not convey precise object information.
We design Language Hierarchical Self-training (LHST) that introduces language hierarchy into weakly-supervised detector training for learning more generalizable detectors.
LHST expands the image-level labels with language hierarchy and enables co-regularization between the expanded labels and self-training. Specifically, the expanded labels regularize self-training by providing richer supervision and mitigating the image-to-box label mismatch, while self-training allows assessing and selecting the expanded labels according to the predicted reliability. 
In addition, we design language hierarchical prompt generation that introduces language hierarchy into prompt generation which helps bridge the vocabulary gaps between training and testing.
Extensive experiments show that the proposed techniques achieve superior generalization performance consistently across 14 widely studied object detection datasets.
\end{abstract}

\section{Introduction}

Object detection aims to locate and identify objects in images by providing basic visual information of ``where and what objects are''. Thanks to the recent advances of deep neural networks, it has achieved great success with various applications in autonomous driving~\cite{cordts2016cityscapes,neuhold2017mapillary,han2021soda10m,zhang2023detr}, intelligent surveillance~\cite{luo2018mio,yongqiang2021baai,wen2020ua,du2018unmanned}, wildlife tracking~\cite{geirdrange_2020,african_animals_dataset,animal-detection-7vafe_dataset}, etc. 
However, learning a generalizable object detector for various downstream tasks that have different data distributions and data vocabularies remains an open research challenge. To this end, weakly-supervised object detection (WSOD)~\cite{bilen2016weakly, redmon2017yolo9000,ramanathan2020dlwl,zhou2022detecting}, which allows access of large-scale image-level datasets (e.g., ImageNet-21K~\cite{deng2009imagenet} with 14M images of 21K classes) with super rich data distributions and data vocabularies, has reignited new research interest under the context of learning generalizable detectors.

While exploiting WSOD to learn generalizable detectors, one typical challenge is that the provided image-level labels do not convey precise object information~\cite{zhou2022detecting} and often mismatch with box-level labels. Recent methods address this challenge by designing various label-to-box assignment strategies that assign the image-level labels to the predicted top-score~\cite{redmon2017yolo9000,ramanathan2020dlwl} or max-size~\cite{zhou2022detecting} object proposals. However, 
the mismatch problem remains due to the restriction of the raw image-level labels~\cite{rasheed2022bridging}. 
At the other end, self-training~\cite{sohn2020simple,zhang2022spectral,huang2021cross} with the detectors pre-trained with~\cite{redmon2017yolo9000,ramanathan2020dlwl,zhou2022detecting} can generate box-level pseudo labels without the restriction of image-level labels. It allows learning from more object proposals without the image-to-box label mismatch issue, but it does not benefit much from the provided image-level label supervision.

\begin{figure*}[t]
\centering
\includegraphics[width=.98\linewidth]{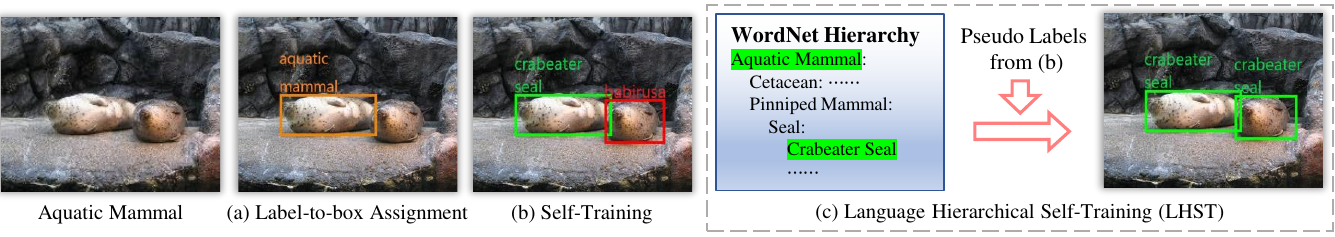}
\caption{
Image-level labels in large-scale datasets such as ImageNet-21k~\cite{deng2009imagenet} often do not convey precise object information~\cite{rasheed2022bridging,zhou2022detecting} which affects while learning generalizable detectors. Recent methods tackle this issue by various label-to-box assignment strategies~\cite{bilen2016weakly,redmon2017yolo9000, ramanathan2020dlwl, zhou2022detecting} as in (a) but are heavily restricted by raw image-level labels and still suffer from image-to-box label mismatch~\cite{rasheed2022bridging}. Self-training~\cite{sohn2020simple} with the detectors pre-trained with~\cite{redmon2017yolo9000,ramanathan2020dlwl,zhou2022detecting} could circumvent the label mismatch issue but the generated pseudo box labels are error-prone due to the lack of proper supervision as in (b). Our proposed LHST introduces language hierarchy to expand the image-level labels and enables co-regularization between the expanded labels and self-training which allows producing more accurate pseudo box labels in (c). 
}
\label{LHST_fig:intro}
\end{figure*}

We propose to incorporate image-level supervision with self-training for learning generalizable detectors, aiming to benefit from self-training while effectively 
making use of
image-level weak supervision. We start from a simple observation: the image-to-box label mismatch largely comes from the ambiguity in language hierarchy, e.g., the image-level label \textit{Aquatic Mammal} in Figure~\ref{LHST_fig:intro} can cover different object-level labels such as seals, dolphins, walruses, etc. 
With the above observations, we design a \textbf{Det}ector with \textbf{L}anguage \textbf{H}ierarchy (DetLH) that combines language hierarchical self-training (LHST) and language hierarchical prompt generation (LHPG) for learning generalizable detectors.

LHST introduces WordNet's language hierarchy~\cite{miller1995wordnet} to expand the image-level labels and accordingly enables co-regularization between the expanded labels and self-training. Specifically, the expanded labels are not all reliable though they can mitigate the image-to-box label mismatch problem by providing richer supervision. Here self-training can predict reliability scores for the expanded labels for better selection or weightage of the expanded labels. At the other end, self-training with pseudo box labels allows learning from more proposals
and can circumvent the image-to-box label mismatch
, but 
the box-level pseudo labels
are usually noisy and may lead to learning degradation~\cite{zhou2022detecting}. Here the expanded labels provide richer and more flexible supervision which can effectively help suppress prediction noises in self-training.

LHPG helps bridge the vocabulary gaps between training and testing by introducing WordNet's language hierarchy into prompt generation process. 
Specifically, LHPG leverages the CLIP language encoder~\cite{radford2021learning} to measure the embedding distances between test concepts and WordNet synsets, and then generates the prompt for a given test concept from its best matched WordNet synset.
In this way, the test prompts generated by LHPG have been standardized by WordNet and are well aligned with our proposed detector that is trained with WordNet information via LHST.
In another word, the combination of LHST and LHPG actually leverages WordNet as a standard and intermediate vocabulary that bridges the gaps between training and testing vocabularies, generating better prompts and leading to better detection performance on downstream applications.

The main contributions of this work are threefold.
\textit{First}, we propose language hierarchical self-training that incorporates language hierarchy with self-training for weakly-supervised object detection.
\textit{Second}, we design language hierarchical prompt generation, which introduces language hierarchy into prompt generation to bridge the vocabulary gaps between detector training and testing.
\textit{Third}, extensive experiments show that 
our DetLH achieves superior generalization performance consistently across 14 detection benchmarks.

\section{Related Work}

\textbf{Weakly-supervised object detection (WSOD)} aims to train object detectors using image-level supervision. Traditional WSOD methods~\cite{li2019weakly,shen2020enabling,shen2019cyclic,wan2019c,yang2019towards} use image-level annotations only without any box annotations and thus focus on low-level proposal mining techniques~\cite{arbelaez2014multiscale,uijlings2013selective,bilen2016weakly,tang2017multiple,tang2018pcl,huang2020comprehensive}, leading to unsatisfying localization performance.
\textbf{Semi-supervised WSOD}~\cite{dong2021boosting,fang2021wssod,li2018mixed,liu2021mixed,uijlings2018revisiting,yan2017weakly,zhong2020boosting} has been proposed to further improve the performance, which leverages both box-level and image-level annotated data.
With better localization quality, recent methods~\cite{redmon2017yolo9000,ramanathan2020dlwl,zhou2022detecting,zhang2021mosaicos} design various label-to-box assignment strategies, such as assigning image-level labels to max-score anchors~\cite{redmon2017yolo9000}, max-score proposals~\cite{ramanathan2020dlwl} or max-size proposals~\cite{zhou2022detecting}.
Our work belongs to semi-supervised WSOD. Different from previous methods, we tackle the image-to-box label mismatch by introducing language hierarchy into self-training.

\textbf{Large-vocabulary object detection}~\cite{gupta2019lvis,redmon2017yolo9000,singh2018r,yang2019detecting,zhang2024vision} researches on detecting thousands of categories. Most previous papers focus on tackling the long-tail issue~\cite{chang2021image,feng2021exploring,li2020overcoming,pan2021model,wu2020forest,zhang2021distribution}, e.g., by using equalization losses~\cite{tan2021equalization,tan2020equalization}, SeeSaw loss~\cite{wang2021seesaw}, or Federated Loss~\cite{zhou2021probabilistic}. Recent semi-supervised WSOD methods~\cite{redmon2017yolo9000,ramanathan2020dlwl,zhou2022detecting} and our work circumvent the long-tail problem by leveraging more balanced image-level datasets such as ImageNet-21K.

\textbf{Open-vocabulary object detection} focuses on detecting objects conditioned on arbitrary words (i.e., any category names). A common strategy~\cite{bansal2018zero,pennington2014glove,rahman2020improved,li2019zero,zareian2021open} is to replace the detector’s classification layer with the language embeddings of category names. 
Recent methods~\cite{gu2021open,ghiasi2021open,li2022language,zang2022open,rasheed2022bridging,zhou2022detecting} leverage the powerful CLIP~\cite{radford2021learning} model by using its text embeddings~\cite{gu2021open,ghiasi2021open,li2022language,zang2022open,rasheed2022bridging,zhou2022detecting} or conducting knowledge distillation~\cite{gu2021open,zang2022open,rasheed2022bridging}.
Similar to Detic~\cite{zhou2022detecting}, our work uses CLIP text embeddings as the classifier and leverages image-level annotated data instead of distilling knowledge from CLIP.

\textbf{Language hierarchy} has been widely studied for visual recognition tasks~\cite{silla2011survey}, especially for large-vocabulary visual recognition.
Most existing studies~\cite{dimitrovski2011hierarchical,liu2021improving,chalkidis2020empirical} focus on image classification tasks, e.g., leveraging language hierarchy for multi-label image classification~\cite{dimitrovski2011hierarchical,liu2021improving,chalkidis2020empirical,chen2021hsva,mensink2014costa,yi2022exploring,cao2020zero}, modelling hierarchical relations among classes~\cite{chen2021hsva,mensink2014costa} or facilitating classification training~\cite{yi2022exploring,cao2020zero}.
Different from previous work, we introduce language hierarchy into self-training for weakly-supervised object detection.

\section{Method}

This work focuses on learning generalizable object detectors via weakly-supervised detector training~\cite{zhou2022detecting}, which leverages additional large-scale image-level datasets to enlarge the data distributions and data vocabularies in detector training. We first describe the task definition with training and evaluation setups. 
Then, we present our proposed DetLH which is detailed in two major aspects on Language Hierarchical Self-training (LHST) that introduces language hierarchy into detector training, and Language Hierarchical Prompt Generation (LHPG) that introduces language hierarchy into prompt generation.

\subsection{Task Definition}
\label{sec31}

\textbf{Training setup.} The training data consists of two parts: 1) a detection dataset $\mathcal{D}_{det} = \{(x, y_{det})_{i}\}_{i=1}^{|{D}_{det}|}$, where $x$ denotes an image while $y_{det}$ stands for the class and bounding box labels for $x$; 2) an image classification dataset $\mathcal{D}_{cls} = \{(x, y_{cls})_{i}\}_{i=1}^{{|{D}_{cls}|}}$ where $y_{cls}$ denotes the image-level label (i.e., a one-hot vector) for $x$.
Given the two datasets, the goal is to learn a generalizable detection model $F$ by jointly optimizing $F$ over $\mathcal{D}_{det}$ and $\mathcal{D}_{cls}$:

\begin{equation}
Loss = \sum_{(x, y_{det}) \in \mathcal{D}_{det}}\mathcal{L}_{det}(F(x), y_{det})
+ \sum_{(x, y_{cls}) \in \mathcal{D}_{cls}}\mathcal{L}_{weak}(F(x), y_{cls}),
\label{eq_baseline_loss}
\end{equation}
where $\mathcal{L}_{det}(\cdot) = \mathcal{L}_{rpn}(\cdot) + \mathcal{L}_{reg}(\cdot) + \mathcal{L}_{cls}(\cdot)$ is the fully-supervised detection loss function while $\mathcal{L}_{rpn}(\cdot)$, $\mathcal{L}_{reg}(\cdot)$, and $\mathcal{L}_{cls}(\cdot)$ denote RPN, Regression, and Classification loss functions, respectively. $\mathcal{L}_{weak}$ is the weakly-supervised loss function to train detectors with image-level labels.

\textbf{Evaluation setup.} As the goal is to learn a generalizable detection model that works well
on various unseen downstream tasks, we conduct zero-shot cross-dataset evaluation\footnote{zero-shot cross-dataset evaluation here means that the model is evaluated on unseen datasets, which is the same as the one defined in CLIP~\cite{radford2021learning}.} to assess the generalization performance of the trained detection model.
Note, different domain adaptation~\cite{wang2018deep,huang2021model,zhang2021prototypical,huang2022category} that generally uses downstream data in training, our setup is similar to domain generalization~\cite{huang2021fsdr,zhou2022domain} that does not involve downstream data in training.

\textbf{Open-vocabulary Detector.} We modify the classification layer of the detector into an open-vocabulary format such that the detector could be tested over unseen datasets.
Specifically, we replace the weights of the detector's classification layer with the fixed language embeddings encoded from class names, where the object classification could be achieved by matching the object's embedding and the fixed language embeddings. We adopt the CLIP language embeddings~\cite{radford2021learning} as the classification weights as in~\cite{zhou2022detecting,gu2021open}. In this way, the modified detector could theoretically detect any target concepts on any target data. 
As reported in~\cite{zhou2022detecting}, detectors trained solely on detection datasets often exhibits constrained performance due to the small-scale training images and vocabularies.
Similar to~\cite{zhou2022detecting}, our proposed DetLH introduces large-scale image-level datasets to enlarge the data distributions and data vocabularies in detector training, leading to more generalizable detectors and better generalization performance on various unseen datasets.

\begin{figure*}[t]
\centering
\includegraphics[width=.98\linewidth]{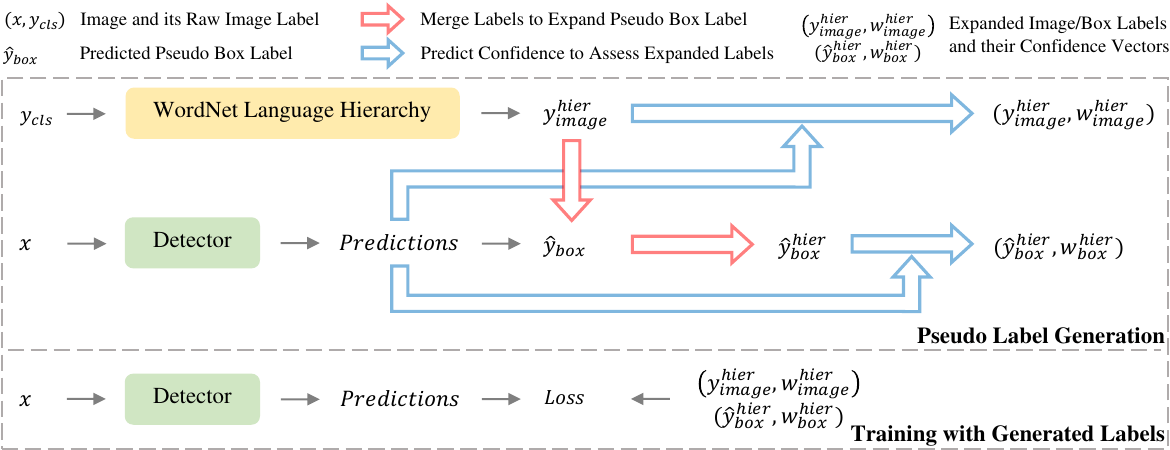}
\caption{The proposed language hierarchical self-training consists of two flows including Pseudo Label Generation (top box) and Training with Generated Labels (bottom box).
The Pseudo Label Generation flow leverages WordNet to expand the image-level labels, and then merges the expanded image-level labels with the predicted pseudo box labels, such that the expanded image-level labels could provide richer and more flexible supervision (than the limited and rigid raw labels) to regularize the self-training 
which is prone to errors in pseudo labeling.
In addition, as the labels expanded by WordNet (i.e., the expanded logits `1' in $y_{image}^{hier}$ and $y_{box}^{hier}$) are not all reliable, Pseudo Label Generation predicts reliability scores for the expanded labels to adaptively re-weight them when applying them on different images or pseudo boxes.
In Training with Generated Labels, we optimize the detector with the generated image-level and box-level labels, where the image-level training
could regularize the training with pseudo box-level labels as pseudo box labels vary along training iterations and are not very stable.
}
\label{LHST_fig:framework}
\end{figure*}

\subsection{Language Hierarchical Self-training}

The proposed LHST utilizes WordNet’s language hierarchy to expand the image-level labels, which enables co-regularization between the expanded image-level labels and self-training as illustrated in Figure~\ref{LHST_fig:framework}.

\textbf{Overview.} For \textit{fully supervised detector training} over the detection dataset, we feed box-level annotated samples $(x, y_{det}) \in \mathcal{D}_{det}$ to the detection model $F$ and optimize $F$ with the standard fully supervised detection loss, i.e., the first term of Eq.~\ref{eq_baseline_loss}. 
For \textit{weakly-supervised detector training} over the image-level annotated dataset $(x, y_{cls}) \in \mathcal{D}_{cls}$ shown in Figure~\ref{LHST_fig:framework}, we first leverage WordNet’s language hierarchy to expand the raw image-level label $y_{cls}$ into $y_{image}^{hier}$ (the hierarchical image-level label), and merge $y_{image}^{hier}$ and the generated pseudo box label $\hat{y}_{box}$ to acquire $\hat{y}_{box}^{hier}$ (the hierarchical box-level pseudo label). Then, we optimize the detector with $(\hat{y}_{box}^{hier}, w_{box}^{hier})$ and $(y_{image}^{hier}, w_{image}^{hier})$, where $w_{image}^{hier}$ and $w_{box}^{hier}$ denote the predicted reliability scores of the expanded logits `1' in $y_{image}^{hier}$ and $y_{box}^{hier}$ and are used to weight the labels in loss calculation.

\textbf{Expanding image labels with language hierarchy.} Given image-level annotated dataset $(x, y_{cls}) \in \mathcal{D}_{cls}$ ($y_{cls}$ is a label vector with length $C$ and $C$ denotes the number of classes), we leverage WordNet's class name hierarchy~\cite{miller1995wordnet} to expand $y_{cls}$ into $y_{image}^{hier}$ as the following:

\begin{equation}
y_{image}^{hier}  = \text{WordNet}(y_{cls}),
\label{wordnet_labeling}
\end{equation}
where the function $\text{WordNet}(\cdot)$ recursively finds all hypernyms and hyponyms of the input (i.e., the class indicated in $y_{cls}$) and sets their positions in the label vector $y_{cls}$ to be `1' to expand $y_{cls}$ into $y_{image}^{hier}$. In this way, a single-label annotation could be expanded into a multi-label annotation within the very rich  ImageNet-21K vocabulary.

\textbf{Generating pseudo box labels with predictions.} Given the image $x \in \mathcal{D}_{cls}$, we feed $x$ into the detector $F$ to acquire the prediction as following:
\begin{equation}
\{p_{n}^{c}\}_{1 \leq n \leq N, 1 \leq c \leq C} = F(x),
\label{pseudo_gen}
\end{equation}
where $p_{n}$ is the probability vector of the predicted $n$-th bounding box after Softmax, and $p_{n}^{c}$ denotes the predicted $c$-th category probability.
Note we filter out a prediction if its max confidence score is lower than the threshold $t$, and $N$ denotes the number of predicted object proposals after filtering, i.e., $max(\{p_{n}^{c}\}_{1\leq c\leq C}) \geq t, \forall n$.

Then the pseudo category label $\hat{y}_{box} = \{\hat{y}_{n}\}_{1 \leq n \leq N}$ for $N$ boxes in image $x$ is derived by:
\begin{equation}
\argmax_{\hat{y}_{n}} \sum_{c = 1}^{C} \hat{y}_{n}^{c} \log p_{n}^{c}, \ s.t. \ \hat{y}_{n} \in \Delta^{C}, \forall n,
\end{equation}
where $\hat{y}_{n} = (\hat{y}^{(1)}_{n}, \hat{y}^{(2)}_{n}, ..., \hat{y}^{(C)}_{n})$ is the predicted category label, and $ \Delta^{C}$ denotes a probability simplex with length $C$.

\textbf{Merging image and pseudo box labels.} As the predicted pseudo box label $\hat{y}_{box}$ is error-prone, we regularize it with the expanded image-level supervision by merging $y_{image}^{hier}$ and $\hat{y}_{box}$ as the following: 
\begin{equation}
\hat{y}_{box}^{hier}(n) =  \hat{y}_{box}(n) \lor y_{image}^{hier}, \forall n,
\end{equation}
where $\lor$ denotes the logical ``OR'' operator.

\textbf{Assessing the expanded labels.} As the labels expanded by WordNet (i.e., the expanded logits `1' in $\hat{y}_{box}^{hier} = \{\hat{y}_{n}^{c}\}_{1 \leq n \leq N, 1 \leq c \leq C}$) are not all reliable, we predict a reliability score $w_{n}^{c}$ for the expanded label to adaptively re-weight $y_{n}^{c} \in \hat{y}_{box}^{hier}$ when applying it on different pseudo boxes. We measure the reliability of $y_{n}^{c}$ with prediction $p_{n}^{c}$, and $w_{box}^{hier} = \{w_{n}^{c}\}_{1 \leq n \leq N, 1 \leq c \leq C}$ can be derived by:

\begin{equation}
  w_{n}^{c} =
  \begin{cases}
    p_{n}^{c} & \text{if ${y_{image}^{hier}}^{(c)} \neq y_{cls}^{(c)}$} \\
    1 & \text{otherwise},
  \end{cases}
\label{eq:weight}
\end{equation}
where ${y_{image}^{hier}}^{(c)} \neq y_{cls}^{(c)}$ returns True if the $c$-th label logit in $y_{image}^{hier}$ is expanded by WordNet, which also applies to $\hat{y}_{box}^{hier}$ as $\hat{y}_{box}^{hier}$ is expanded by mergeing it with $y_{image}^{hier}$.

Given the prediction $\{p_{n}^{c}\}_{1 \leq n \leq N, 1 \leq c \leq C}$, the merged pseudo box label $\hat{y}_{box}^{hier} = \{\hat{y}_{n}^{c}\}_{1 \leq n \leq N, 1 \leq c \leq C}$ and its reliability score $w_{box}^{hier} = \{w_{n}^{c}\}_{1 \leq n \leq N, 1 \leq c \leq C}$, we optimize the detector $F$ as the following:
\begin{equation}
\mathcal{L}_{box}(F(x)) = \sum_{n}^{N}\sum_{c}^{C} (\text{BCE}(p_{n}^{c},y_{n}^{c}) \times w_{n}^{c}),
\end{equation}
where $\text{BCE}(\cdot)$ denotes the binary cross-entropy loss.

In addition, training with the predicted pseudo box labels is not very stable as pseudo box labels vary along training process. Thus, we regularize the training of $\mathcal{L}_{box}(F(x))$ with an image-level loss defined as the following:
\begin{equation}
\mathcal{L}_{image}(F(x)) = \sum_{c}^{C} (\text{BCE}(p^{c}_{image}, {y_{image}^{hier}}^{(c)}) \times w^{c}_{image}),
\end{equation}where $p_{image} = \{p_{image}^{c}\}_{1 \leq c \leq C}$ denotes the category probability predicted for the image-level proposal. $w_{image}^{hier} = \{w^{c}_{image}\}_{1 \leq c \leq C}$ denotes the reliability score for the expanded logits ``1" in $y_{image}^{hier}$. Similar to Eq.~\ref{eq:weight}, $w^{c}_{image} = p_{image}^{c}$ if ${y_{image}^{hier}}^{(c)} \neq y_{cls}^{(c)}$, otherwise $w^{c}_{image} = 1$. Besides, $\text{BCE}(\cdot)$ denotes the binary cross-entropy loss.

\textbf{Training objective.} The overall training objective of Language Hierarchical Self-training is defined as:
\begin{equation}
\mathcal{L}_{lhst} = \sum_{(x, y_{det}) \in \mathcal{D}_{det}}\mathcal{L}_{det}(F(x), y_{det})
+ \sum_{(x, y_{cls}) \in \mathcal{D}_{cls}}(\mathcal{L}_{box}(F(x)) + \mathcal{L}_{image}(F(x)) )
\end{equation}

\textbf{Language Hierarchical Prompt Generation.}
As the goal is to learn a generalizable detection model that works well on various downstream tasks, one typical challenge is the vocabulary gap between detector training datasets (i.e., LVIS and ImageNet-21K) and detector testing datasets (e.g., object365 or customized data). 
A common solution of tackling the vocabulary gaps is to conduct prompt learning~\cite{zhou2022conditional} to generate proper category prompts. 
However, prompt learning generally requires labeled target images for additional training.

In this work, we tackle the vocabulary gaps by generating prompts with the help of WordNet, which introduces little computation overhead and does not require labeled target images and additional training.
To this end, we design language hierarchical prompt generation (LHPG) that works by incorporating WordNet information into prompt generation process.
Specifically, LHPG leverages CLIP language encoder~\cite{radford2021learning} to measure the embedding distances between test concepts and WordNet synsets, and then generates the prompt for a given test concept from its best matched WordNet synset: $ V_{test}^{\text{WordNet}} = \text{CLIP}(V_{test}, \text{WordNet})$, where $V_{test}$ denotes test vocabulary, $\text{WordNet}$ denotes WordNet synsets, $\text{CLIP}$ denotes CLIP language encoder and $V_{test}^{\text{WordNet}}$ stands for the best matched WordNet synsets for the classes in $V_{test}$.
Then, we generate test prompts from $V_{test}^{\text{WordNet}}$.
As compared with $V_{test}$, our $V_{test}^{\text{WordNet}}$ has been standardized by WordNet and is well aligned with our proposed detector that is 
trained with WordNet information via LHST.
In another word, the combination of LHST and LHPG makes use of WordNet as a standard and intermediate vocabulary that bridges the gaps between training and testing vocabularies, generating better prompts and leading to better detection performance on downstream applications.

\section{Experiments}
We evaluate our DetLH 
on 14 widely adopted detection benchmarks. We follow the zero-shot cross-dataset object detection setting proposed in ~\cite{rasheed2022bridging,zhou2022detecting}.
More details like \textbf{Dataset} and \textbf{Implementation Details} are provided in the appendix.

\begin{table*}[ht]
\centering
\caption{
\textbf{Zero-shot cross-dataset object detection for common objects.
} 
All detectors are trained over the training datasets (LVIS and ImageNet-21K) and evaluated over target datasets (i.e., Object365 and Pascal VOC with objects from common classes and scenarios) without finetuning. 
``Dataset-specific oracles'' denote the detectors that are 
fully supervised which are trained by using the training data of respective datasets.
}
\label{tab:general}
\resizebox{\linewidth}{!}{
\begin{tabular}{lcccccc|cccccc}
\toprule
\multirow{2}{*}{Method} &\multicolumn{6}{c}{Object365 \cite{shao2019objects365}} &\multicolumn{6}{c}{Pascal VOC \cite{everingham2009pascal}}\\\cmidrule{2-13}
 & AP & AP50 & AP75 &APs &APm &APl & AP & AP50 & AP75 &APs &APm &APl\\ \midrule
WSDDN~\cite{bilen2016weakly} &21.0 &29.1 &22.7 &8.7 &20.9 &31.2 &61.6 &82.7 &67.5 &24.8 &50.9 &73.5\\
YOLO9000~\cite{redmon2017yolo9000} &21.0 &28.5 &22.6 &8.6 &20.7 &30.9 &62.6 &83.6 &68.7 &23.7 &52.0 &73.9\\
DLWL~\cite{ramanathan2020dlwl} &21.3 &29.1 &23.0 &8.8 &21.0 &31.5 &62.4 &83.4 &68.3 &23.8 &51.2 &73.8\\
Detic~\cite{zhou2022detecting} &21.6 &29.4 &23.4 &9.0 &21.4 &31.9 &62.4 &83.3 &68.5 &23.7 &51.8 &73.9\\
\textbf{DetLH (Ours)} &\textbf{23.6} &\textbf{32.5} &\textbf{25.5} &\textbf{9.8} &\textbf{23.5} &\textbf{35.0} &\textbf{64.4} &\textbf{86.1} 
 &\textbf{70.8} &\textbf{25.3} &\textbf{54.1} &\textbf{75.3}\\
\midrule
Dataset-specific oracles &31.2 &- &-&-&-&-&54.4 &79.7	&59.1	&19.0 &40.8 &64.5
\\
\bottomrule
\end{tabular}
}
\end{table*}

\begin{table*}[ht]
\centering
\caption{
\textbf{Zero-shot cross-dataset object detection for autonomous driving.
} 
All detectors are trained over the training datasets (LVIS and ImageNet-21K) and evaluated over autonomous driving datasets (i.e., Cityscapes, Vistas and SODA10M) without finetuning. 
}
\label{tab:auto}
\resizebox{\linewidth}{!}{
\begin{tabular}{lccc|ccc|ccc|ccc}
\toprule
\multirow{2}{*}{Method} &\multicolumn{3}{c}{Cityscapes \cite{cordts2016cityscapes}} &\multicolumn{3}{c}{Vistas \cite{neuhold2017mapillary}} &\multicolumn{3}{c}{SODA10M \cite{han2021soda10m}} & \multicolumn{3}{c}{Average}\\ \cmidrule{2-13}
 & AP & AP50 & AP75  & AP & AP50 & AP75 & AP & AP50 & AP75 & AP & AP50 & AP75 \\ \midrule
WSDDN~\cite{bilen2016weakly} &28.2 &45.4 &27.1 &22.3 &34.0 &23.3 &17.4 &28.9 &17.1 &22.6 &36.1 &22.4 \\
YOLO9000~\cite{redmon2017yolo9000} &28.8 &46.2 &27.4 &22.5 &34.6 &23.4 &18.3 &30.4 &18.0 &23.2 &37.0 &22.9 \\
DLWL~\cite{ramanathan2020dlwl} &28.6 &45.6 &28.1 &22.5 &34.7 &23.2 &18.3 &30.4 &18.0 &23.1 &36.9 &23.0 \\
Detic~\cite{zhou2022detecting} & 29.6 &47.1 &28.4 &23.0	&35.6	&23.6 &18.8 & 30.9 & 18.5 &23.8 &37.9 &23.5 \\
\textbf{DetLH (Ours)} &\textbf{31.2}	&\textbf{50.3} &\textbf{29.1} &\textbf{26.5} &\textbf{44.0} &\textbf{25.8} &\textbf{25.1}	&\textbf{38.4} &\textbf{26.1} &\textbf{27.6} &\textbf{44.2} &\textbf{27.0} 
\\
\midrule
Dataset-specific oracles &43.0	&69.0	&42.6 &28.1	&45.8	&28.5 &44.7	&68.2	&47.3 &38.6 &61.0 &39.5 
\\
\bottomrule
\end{tabular}
}
\end{table*}

\begin{table*}[ht]
\centering
\caption{
\textbf{Zero-shot cross-dataset object detection under different weather and time-of-day conditions (using metric AP50).
} 
All detectors are trained over the training datasets (LVIS and ImageNet-21K) and evaluated over BDD100K and DAWN datasets without finetuning. 
}
\label{tab:weather}
\resizebox{\linewidth}{!}{
\begin{tabular}{lcccccc|cccc|ccc|ccc}
\toprule
\multirow{2}{*}{Method} &\multicolumn{6}{c}{BDD100K-weather \cite{yu2020bdd100k}} &\multicolumn{4}{c}{BDD100K-time-of-day \cite{yu2020bdd100k}} & \multicolumn{3}{c}{DAWN \cite{kenk2020dawn}} & \multicolumn{1}{c}{Avg}\\ \cmidrule{2-15}
& rainy &snowy &overcast &cloudy &foggy &undefined &daytime &dawn\&dusk &night &undefined &fog &sand &snow \\ \midrule
WSDDN~\cite{bilen2016weakly} &35.0 &33.0 &38.3 &41.7 &26.7 &46.0 &39.1 &35.5 &27.9 &50.2 &62.6 &55.0 &65.6 &42.8 \\
YOLO9000~\cite{redmon2017yolo9000}&34.4 &33.6 &39.5 &41.8 &31.0 &45.4 &39.6 &35.9 &28.8 &46.6 &60.6 &53.9 &64.4 &42.7  \\
DLWL~\cite{ramanathan2020dlwl} &34.8 &33.4 &38.8 &43.8 &\textbf{40.2} &45.2 &40.1 &35.1 &28.7 &45.0 &62.1 &56.1 &63.7 &43.6 \\
Detic~\cite{zhou2022detecting} &34.3 &33.2 &39.5 &41.9 &27.9 &45.4 &39.2 &35.5 &28.8
&48.2 &52.3 &54.1 &56.1 &41.3 \\
\textbf{DetLH (Ours)} &\textbf{40.2} &\textbf{37.5} &\textbf{48.2} &\textbf{49.3} &37.1 &\textbf{49.9} &\textbf{45.7} &\textbf{40.0} &\textbf{34.2} &\textbf{53.0} &\textbf{63.2} &\textbf{57.6} &\textbf{67.3} &\textbf{47.9} \\
\midrule
Dataset-specific oracles &52.0 &52.5 &56.3 &56.3 &21.3 &65.4 & 57.0 & 50.4 &48.6 & 27.7 &56.7 &48.4 &26.4 & 47.6 
\\
\bottomrule
\end{tabular}
}
\end{table*}

\begin{table*}[!ht]
\centering
\caption{
\textbf{Zero-shot cross-dataset object detection for intelligent surveillance.
} 
All detectors are trained over the training datasets (LVIS and ImageNet-21K) and evaluated over surveillance datasets MIO-TCD, BAAI-VANJEE, DETRAC and UAVDT without finetuning. 
}
\label{tab:surve}
\resizebox{\linewidth}{!}{
\begin{tabular}{lccc|ccc|ccc|ccc|ccc}
\toprule
\multirow{2}{*}{Method} &\multicolumn{3}{c}{MIO-TCD \cite{luo2018mio}} &\multicolumn{3}{c}{BAAI-VANJEE \cite{yongqiang2021baai}} &\multicolumn{3}{c}{DETRAC \cite{wen2020ua}} &\multicolumn{3}{c}{UAVDT \cite{du2018unmanned}} & \multicolumn{3}{c}{Average}\\ \cmidrule{2-16}
& AP & AP50 & AP75 & AP & AP50 & AP75 & AP & AP50 & AP75 & AP & AP50 & AP75 & AP & AP50 & AP75  \\ \midrule
WSDDN~\cite{bilen2016weakly} &11.3 &17.6 &11.6 &13.1 &19.6 &13.3 &25.6 &35.3 &31.1 &17.1 &31.9 &16.0 &16.7 &26.1 &18.0\\
YOLO9000~\cite{redmon2017yolo9000} &12.7 &19.7 &13.0 &13.1 &19.3 &13.0 &29.1 &39.4 &35.3 &18.6 &33.9 &17.7 &18.3 &28.1 &19.7 \\
DLWL~\cite{ramanathan2020dlwl} &12.9 &20.1 &12.9 &13.5 &20.0 &13.6 &27.8 &38.0 &33.6 &16.6 &31.1 &15.1 &16.7 &26.1 &18.0 \\
Detic~\cite{zhou2022detecting} &13.4 &20.6	&13.9	&16.9	&23.6	&17.6	&28.7	&39.2	&34.8	&18.6	&34.2	&17.6 &19.4 &29.4 &21.0 \\
\textbf{DetLH (Ours)} &\textbf{15.8}	&\textbf{24.5}	&\textbf{16.0} &\textbf{17.9}	&\textbf{25.1}	&\textbf{18.5} &\textbf{32.7}	&\textbf{44.0}	&\textbf{39.7} &\textbf{20.1} &\textbf{36.6} &\textbf{19.3} &\textbf{21.6} &\textbf{32.6} &\textbf{23.4}  \\
\midrule
Dataset-specific oracles &45.2	&63.1	&50.8 &40.6	&58.6	&43.7 &53.1	&70.6	&63.5 &33.8	&60.4	&35.2 &43.2 &63.2 &48.3 
\\
\bottomrule
\end{tabular}
}
\end{table*}

\begin{table*}[ht]
\centering
\caption{
\textbf{Zero-shot cross-dataset object detection for Wildlife Detection.
} 
All detectors are trained over the training datasets (LVIS and ImageNet-21K) and evaluated over wildlife datasets (i.e., Arthropod Detection, AfricanWildlife and Animals Detection) without finetuning. 
}
\label{tab:animal}
\resizebox{\linewidth}{!}{
\begin{tabular}{lccc|ccc|ccc|ccc}
\toprule
\multirow{2}{*}{Method} &\multicolumn{3}{c}{Arthropod Detection \cite{geirdrange_2020}} &\multicolumn{3}{c}{AfricanWildlife \cite{african_animals_dataset}} &\multicolumn{3}{c}{Animals Detection \cite{animal-detection-7vafe_dataset}} & \multicolumn{3}{c}{Average}\\ \cmidrule{2-13}
& AP & AP50 & AP75 & AP & AP50 & AP75 & AP & AP50 & AP75 & AP & AP50 & AP75\\ \midrule

WSDDN~\cite{bilen2016weakly} &18.1 &26.3 &18.8 &\textbf{76.7} &\textbf{88.2} &\textbf{84.0} &36.0 &41.7 &37.5 &43.6 &52.0 &46.7 \\
YOLO9000~\cite{redmon2017yolo9000}&22.6 &33.2 &22.5 &75.9 &87.5 &83.2 &39.0 &45.4 &40.8 &45.8 &55.3 &48.8 \\
DLWL~\cite{ramanathan2020dlwl} &25.3 &34.7 &26.3 &74.7 &86.2 &81.3 &41.7 &48.1 &43.7 &47.2 &56.3 &50.4 \\
Detic~\cite{zhou2022detecting} &27.4 &36.7 &29.2 &68.9	&80.9 &76.4 &41.1 &47.7 &42.9 &45.8 &55.1 &49.5 \\
\textbf{DetLH (Ours)} &\textbf{36.2} &\textbf{49.0} &\textbf{38.3} &74.8 &87.2 &81.8 &\textbf{44.3} &\textbf{51.2} &\textbf{46.3} &\textbf{51.8} &\textbf{62.5} &\textbf{55.5} \\
\midrule 
Dataset-specific oracles &75.1 &86.3 &79.9 &82.7 &90.9	&89.1 &64.4 &74.6	&69.4 &74.1 &83.9 &79.5 
\\
\bottomrule
\end{tabular}
}
\end{table*}

\subsection{Comparison with the state-of-the-art}

We conduct extensive experiments to benchmark our proposed DetLH with state-of-the-art methods. We evaluate them on 14 widely studied object detection datasets to assess their zero-shot cross-dataset generalization ability.
Tables~\ref{tab:general}-~\ref{tab:animal} report zero-shot cross-dataset detection results for common objects, autonomous driving, intelligent surveillance, and wildlife detection, respectively. More details are to be described in the following paragraphs.

\textbf{Object detection for common objects.} Table~\ref{tab:general} shows that DetLH outperforms state-of-the-art methods clearly on common object datasets Object365 and Pascal VOC. In addition, we can observe that DetLH even brings significant gains above the \textit{dataset-specific oracle} (i.e., the model that is fully trained on the target training data) on Pascal VOC (i.e., a small-scale dataset), showing the advantages of leveraging large-scale training data.

\textbf{Object detection for autonomous driving.} As shown in Table~\ref{tab:auto}, our DetLH outperforms state-of-the-art methods by large margins on various autonomous driving datasets, showing that DetLH still works effectively while facing large variations in camera views from autonomous driving scenarios to the base-dataset scenarios (LVIS and ImageNet-21K), e.g., autonomous driving images are captured under very different camera views.
In addition, the experimental results in Table~\ref{tab:weather} show that our DetLH brings significant performance gains against state-of-the-art methods when encountering various weather and time-of-day conditions, which demonstrates the effectiveness of DetLH while detecting objects under large noises~\cite{huang2021rda}, e.g., the images captured under different weather and  time-of-day conditions may have very different styles and image quality.

\textbf{Object detection for intelligent surveillance.} From Table~\ref{tab:surve}, we can observe that our DetLH outperforms state-of-the-art methods by clear margins on various intelligent surveillance datasets, indicating that DetLH is also tolerant to large changes in the camera lens and angles which often happen to intelligent-surveillance images that are captured under very different camera lens and angles (e.g., surveillance cameras are often with the wide-angle lens and used in high angle views).

\textbf{Object detection for Wildlife.} The experimental results in Table~\ref{tab:animal} show that our DetLH performs well on various wildlife detection datasets, showing that DetLH works effectively for detecting fine-grained categories that exist widely in wildlife detection datasets. The significant performance gains largely come from the 
introduction of language hierarchy into detector training and prompt generation, which helps model the hierarchical relations among parent and fine-grained subcategories and thus leads to better fine-grained object detection.

The superior detection performance of our DetLH is largely attributed to our two core designs, i.e., LHST and LHPG. LHST enables effective usage of large-scale image-level annotated images and significantly enlarges the data distribution and the data vocabulary in detector training, yielding robust performance under large cross-dataset gaps in data distribution and vocabulary.
LHPG ingeniously helps mitigate the vocabulary gaps between detector training and testing. It improves the overall confidence of detection and benefits the detection as large data distribution gaps (or large data vocabulary gaps) often lead to low-confidence predictions and poor detection results.

\subsection{Ablation Studies}
\label{exp:alb}

\begin{wraptable}{r}{6.8cm}
\vspace{-3ex}
\caption{
\small	
\textbf{Ablation studies of our DetLH}
with Language Hierarchical
Self-training (LHST) and Language Hierarchical Prompt Generation (LHPG). 
The experiments are conducted with Swin-B based CenterNet2~\cite{zhou2022detecting} and the detectors are evaluated on Object365 in zero-shot cross-dataset object detection setup.
}
\small
\setlength\tabcolsep{3pt}
\resizebox{6.8cm}{!}{
\centering	
\begin{tabular}{lccc}
\toprule
Method & LHST
& LHPG & AP50\\
\midrule
Box-Supervised~\cite{zhou2022detecting} &&&26.5 \\\midrule
&\checkmark &&31.3\\
&&\checkmark&31.0\\
\textbf{DetLH (Ours)} &\checkmark&\checkmark&\textbf{32.5} \\\bottomrule
\end{tabular}
}
\vspace{-2ex}
\label{tab:abla_o365}
\end{wraptable}	

We perform ablation studies with Swin-B~\cite{liu2021swin} based CenterNet2~\cite{zhou2021probabilistic} over the large-scale Object365 dataset as shown in Table~\ref{tab:abla_o365}. As the core of our proposed DetLH, we examine how our designed LHST and LHPG contribute to the overall performance of zero-shot cross-dataset object detection. As shown in Table~\ref{tab:abla_o365}, the \textit{baseline} (Box-Supervised~\cite{zhou2022detecting}) does not perform well as it uses box-level training data only.
It can be observed that LHST outperforms the baseline clearly, showing that LHST can effectively leverage the large-scale image-level annotated dataset to significantly enlarge the data distribution and data vocabulary involved in detector training, leading to much better zero-shot cross-dataset detection performance.
In addition, LHPG brings clear performance improvements in zero-shot cross-dataset detection by introducing language hierarchy into prompt generation, demonstrating the effectiveness of LHPG in mitigating the vocabulary gaps between training and testing.
Moreover, 
the inclusion of both LHST and LHPG in the proposed DetLH performs clearly the best, indicating the complementary property of our two designs.

\subsection{Discussion}
\label{exp_discussion}

\begin{table}[ht]
\centering
\caption{
\textbf{Zero-shot cross-dataset object detection on various datasets.} Results are averaged on 14 widely studied datasets.
}
\small
\setlength\tabcolsep{3pt}
\resizebox{0.55\linewidth}{!}{
\centering	
\begin{tabular}{lcccccc}
\toprule
\multirow{2}{*}{Method} &\multicolumn{6}{c}{Averaged over 14 detection datasets}\\ \cmidrule{2-7}
 & AP & AP50 & AP75 &APs &APm &APl\\\midrule
WSDDN~\cite{bilen2016weakly}  &29.9 &42.5 &31.4 &14.8 &25.9 &44.2
\\
YOLO9000~\cite{redmon2017yolo9000} &30.9 &43.8 &32.4 &14.1 &25.8 &45.1
\\
DLWL~\cite{ramanathan2020dlwl}  &31.0 &44.0 &32.5 &15.4 &26.3 &45.3
\\
Detic~\cite{zhou2022detecting} &31.0 &44.0 &32.8 &14.6 &27.5 &45.5
\\
\textbf{DetLH (Ours)} &\textbf{34.6} &\textbf{49.3} &\textbf{36.4} &\textbf{16.0} &\textbf{28.4} &\textbf{49.5}
\\
\bottomrule
\end{tabular}
}
\vspace{-2ex}
\label{tab:sum}
\end{table}

\textbf{Generalization across various detection tasks:} We study the generalization of our DetLH by conducting zero-shot cross-dataset object detection on 14 widely studied object detection datasets. Tables~\ref{tab:general}-~\ref{tab:animal} show that DetLH achieves superior performance consistently across all the detection applications. Besides, Table~\ref{tab:sum} summarizes the detection results averaged on 14 datasets, showing that DetLH clearly outperforms the state-of-the-art methods.

\textbf{Generalization across various network architectures:} We study the generalization of the proposed DetLH from the perspective of network architectures. Specifically, we perform extensive evaluations with four representative network architectures, including one Transformer-based (i.e., Swin-B) and three CNN-based (i.e., ConvNeXt-T, ResNet-50 and ResNet-18).
Experimental results in Table~\ref{tab:archi} show that the proposed DetLH outperforms the state-of-the-art method consistently over different network architectures.

\begin{table}[ht]
\centering
\caption{ 
\textbf{Zero-shot cross-dataset object detection with different network architectures.
} 
All networks architectures are trained over the training datasets (LVIS and ImageNet-21K) and evaluated over Object365 without finetuning.
}
\label{tab:archi}
\resizebox{0.8\linewidth}{!}{
\begin{tabular}{llcccccc}
\toprule
\multirow{2}{*}{Method} &\multirow{2}{*}{Architecture} &\multicolumn{6}{c}{Object365}\\ \cmidrule{3-8}
 && AP & AP50 & AP75 &APs &APm &APl\\\midrule
Detic~\cite{zhou2022detecting} &\multirow{2}{*}{Swin-B~\cite{liu2021swin}} &21.6 &29.4 &23.4 &9.0 &21.4 &31.9\\
\textbf{DetLH (Ours)} & &23.6 &32.5 &25.5 &9.8 &23.5 &35.0 \\
\midrule
Detic~\cite{zhou2022detecting} &\multirow{2}{*}{ConvNeXt-T~\cite{liu2022convnet}} &16.9 &23.5 &18.1 &6.8 &16.6 &24.9\\
\textbf{DetLH (Ours)} & &18.9 &26.8 &20.2 &7.6 &18.8 &28.2\\
\midrule
Detic~\cite{zhou2022detecting} &\multirow{2}{*}{ResNet-50~\cite{he2016deep}} &16.2 &22.8 &17.5 &6.3 &16.2 &24.1\\
\textbf{DetLH (Ours)} & &17.7 &25.5 &19.0 &6.9 &17.9 &26.4\\
\midrule
Detic~\cite{zhou2022detecting} &\multirow{2}{*}{ResNet-18~\cite{he2016deep}} &10.8 &15.5 &11.6 &3.9 &10.2 &16.2\\
\textbf{DetLH (Ours)} & &11.8 &17.3 &12.5 &4.3 &11.4 &17.7\\
\bottomrule
\end{tabular}
}
\end{table}

\textbf{Parameter Studies for Language Hierarchical Self-
training (LHST).} In generating pseudo box labels in LHST, we filter out a prediction if its max confidence score is lower than the threshold $t$. We study the threshold $t$ by changing it from $0.65$ to $0.85$ with a step of $0.05$.
Table~\ref{tab:ablation-thres} reports the experimental results on zero-shot transfer object detection over object365 dataset. We can observe that the detection performance is not sensitive to the threshold $t$.

\begin{table}[h]
\caption{
\textbf{Parameter Studies for Language Hierarchical Self-
training (LHST)} on zero-shot transfer object detection over object365 dataset. We study the thresholding parameter $t$ used in generating pseudo box labels in LHST.
}
\begin{center}
\resizebox{0.48\linewidth}{!}{
\begin{tabular}{lccccc}
\toprule
Threshold $t$ &0.65 & 0.70 &0.75 &0.80 &0.85\\
\midrule
AP50 &31.1 &31.3 &31.3 &31.3 &31.2\\
\bottomrule
\end{tabular}
}
\vspace{-3mm}
\end{center}
\label{tab:ablation-thres}
\end{table}

Due to the space limit, we provide more DetLH discussions and visualizations in the appendix.

\section{Conclusion}

This paper presents DetLH, a Detector with Language Hierarchy that combines language hierarchical self-training (LHST) and language hierarchical prompt generation (LHPG) for learning generalizable object detectors.
LHST introduces WordNet’s language hierarchy to expand the image-level labels and accordingly enables co-regularization between the expanded labels and self-training.
LHPG helps mitigate the vocabulary gaps between training and testing by introducing WordNet’s language hierarchy into prompt generation.
Extensive experiments over multiple object detection tasks show that our DetLH achieves superior performance as compared with state-of-the-art methods. In addition, we demonstrate that DetLH works well with different network architectures such as Swin-B, ConvNeXt-T, ResNet-50, etc.
Moving forward, we will explore language hierarchy to further expand the labels in an open-vocabulary manner in addition to the closed ImageNet-21K's vocabulary.

\bibliographystyle{unsrt}
\small\bibliography{main}

\end{document}